\documentclass[letterpaper]{article} 
\usepackage{aaai22} 
\usepackage{times} 
\usepackage{helvet} 
\usepackage{courier} 
\usepackage[hyphens]{url} 
\usepackage{graphicx} 
\usepackage{natbib} 
\usepackage{caption} 
\usepackage[american]{babel}

\usepackage{mathtools} 
\usepackage{siunitx} 
\usepackage{booktabs} 
\usepackage{tikz} 

\usepackage{amsmath,amssymb,amsthm}
\usepackage{enumerate}
\usepackage{booktabs}
\usepackage[switch]{lineno}
\usepackage{xcolor}
\usepackage{xspace}

\usepackage{ifthen}
\newboolean{proofs}
\setboolean{proofs}{true} 
\newcommand{\Proof}[1]{\ifthenelse{\boolean{proofs}}{\begin{proof}\color{green!50!black} #1 \end{proof}}{}}
\newcommand{\RProof}[1]{\ifthenelse{\boolean{proofs}}{\begin{proof}[\textcolor{magenta}{Proof}]\color{magenta} #1 \end{proof}}{}}

\usepackage{relsize}
\usepackage{listings}
\colorlet{darkgreen}{green!80!black}
\colorlet{darkblue}{blue!80!black}
\colorlet{darkred}{red!80!black}

\usepackage[leftmargin=1em]{quoting}

\newcommand{\Omit}[1]{}

\newcommand{\tup}[1]{\langle #1 \rangle}

\newenvironment{example-no-eob}{\noindent\textbf{Example.}\,}{}

\newcommand{\C}{\mathcal{C}}

\newcommand{\Q}{\mathcal{Q}}

\newcommand{\pplus}{\hspace{-.05em}\raisebox{.15ex}{\footnotesize$\uparrow$}}
\newcommand{\mminus}{\hspace{-.05em}\raisebox{.15ex}{\footnotesize$\downarrow$}}

\newcommand{\EQ}[1]{#1{\,=\,}0}
\newcommand{\GT}[1]{#1{\,>\,}0}

\newcommand{\DEC}[1]{#1\mminus}

\newcommand{\UNK}[1]{#1?}

\newcommand{\prule}[2]{\{ #1 \} \mapsto \{ #2 \}}

\newcommand{\siwR}{\ensuremath{\text{SIW}_{R}}\xspace}

\newcommand{\bi}{\begin{itemize}}
\newcommand{\ei}{\end{itemize}}
\newcommand{\bii}{\begin{itemize}}
\newcommand{\eii}{\end{itemize}}

\title{Language-Based Causal Representation Learning}

\author{%
  Blai Bonet,\textsuperscript{\rm 1} \ 
  Hector Geffner\textsuperscript{\rm 2,3}\\
}

\affiliations {
  \textsuperscript{\rm 1}\,Universitat Pompeu Fabra, Barcelona, Spain \\
  \textsuperscript{\rm 2}\,ICREA \& Universitat Pompeu Fabra, Barcelona, Spain \\
  \textsuperscript{\rm 3}\,Link\"{o}ping University, Sweden\\
  bonetblai@gmail.com, hector.geffner@upf.edu
}

\newcommand{\Pick}{\textit{Pick}\xspace}
\newcommand{\Drop}{\textit{Drop}\xspace}
\newcommand{\Move}{\textit{Move}\xspace}

\makeatletter
\def\blfootnote{\gdef\@thefnmark{*}\@footnotetext}
\makeatother

\begin{document}
\allowdisplaybreaks
\maketitle

\begin{abstract}
  Consider the finite state graph that results from a simple, discrete, dynamical system
  in which  an agent  moves in a rectangular grid 
  picking up and dropping packages.
  Can the state variables of the problem, namely, the agent location and the package
  locations, be recovered from the structure of the state graph alone
  without having access to information about the objects, the structure of the states, or any background knowledge?
  We show that this is possible provided that the dynamics is learned over a suitable
  domain-independent \emph{first-order causal language} that makes room for objects
  and relations that are not assumed to be known. The preference for the
  most compact representation in the language that is  compatible with the data provides
  a strong and meaningful learning bias that makes this possible.
  The language of structured causal models (SCMs) is the standard language for representing (static) causal models
  but in dynamic worlds populated by objects, first-order causal languages such as those used
  in ``classical AI planning'' are required. While ``classical AI'' requires handcrafted representations,
  similar representations can be learned from unstructured data over the same languages.
  Indeed, it is the languages and the preference for compact representations in those languages
  that provide structure to the world,  uncovering  objects, relations, and causes.$^*$\blfootnote{The paper was written for a workshop on causal representation learning.
    A valid criticism of ``classical AI'', mentioned in the call, from the perspective of the
    ``new AI'' based on deep learning, is that representations should be learned and not handcrafted.
    One of the aims of the paper was to distinguish representation \emph{languages} from the actual
    representations that they support, as it is only the latter that should be learned. Another aim
    was to emphasize that structural causal models (SCMs) provide one language to talk about causal
    models but not the only one. Action and planning languages have been used to describe causal
    models in AI for a long time too. Our attempt, however, was not successful, as the paper was
    rejected. Not clear if reviewers agreed with these two premises, or found them irrelevant or
    vacuous. For us, the two premises are direct but their implications, explored in this and other
    papers, are not.}
\end{abstract}

\section{Introduction}

Two important challenges in causal representation learning
are learning the state variables of a dynamical system from unstructured data, and learning a representation of the
dynamics that is general and reusable \citep{bernhard:causal,bernhard:causal2}.
For example, a system may involve an agent moving in a $n{\times}m$ grid,
picking up and dropping packages.
Learning the causal structure of the domain means to learn the structure of the
states, given by the agent and the package locations, and the structure of the actions, 
so that they can be used to plan in other instances of this general domain.  
The question is what are the ideas and principles that are required for uncovering
these structures in a crisp and well-founded manner, without involving prior domain knowledge.

For addressing this and other challenges, deep learning approaches usually follow a methodology that goes from
intuitions about inductive biases to deep learning architectures and loss functions, and from there to experimental results and comparisons with
baselines \citep{bengio:high-level,rim,scoff,bengio:rules}. The methodology can be applied broadly and the results show experimental gains, yet the
understanding that follows from them is not always crisp.

In this paper, we articulate a different approach for learning \emph{general causal representations},
and two other representations that exploit causal representations: {general policies}, and {subgoals} (``intrinsic rewards'').
The idea is to learn the representations over suitable \emph{domain-independent languages}
with a known structure (syntax) and known semantics, but without relying on prior knowledge.

For learning representations of general deterministic, discrete dynamics, we appeal to
a language that has been in use in ``classical AI'' since the early 70s; namely, lifted (first-order) STRIPS,
in its modern version, where a planning domain is expressed by a number of \emph{action schemas}
with preconditions and effects given by logical atoms that encode the state variables and their values. 
In ``classical AI'', these action schemas are crafted by hand, but as it has been pointed out by \citet{bernhard:causal2}, this approach
does not scale up as modeling is hard. The ``classical'' approach  does not explain where models come from either.  

\begin{figure*}[t]
  \centering
  \begin{tabular}{c}
    \resizebox{\textwidth}{!}{
      \includegraphics{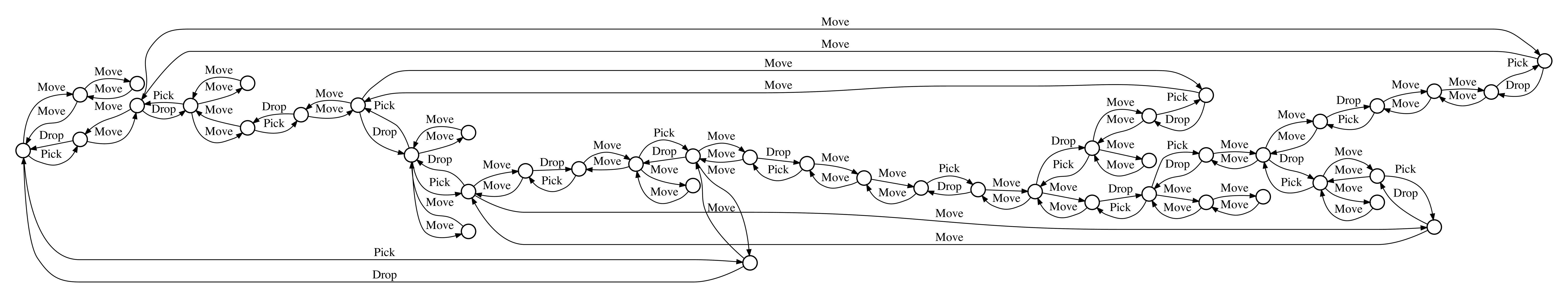}
    } \\[1em]
    \begin{tabular}{ccc}
      \begin{minipage}{.90\columnwidth}
        \begin{lstlisting}[escapechar=|,basicstyle=\ttfamily\scriptsize]
  |\textbf{Move}|(?to,|\,|?from):
    |\textcolor{darkgreen}{Static Pre:}| neq(?to,|\,|?from), p5(?to,|\,|?from)
    |\textcolor{darkblue}{Pre:}|  p2(?from), -p2(?to)
    |\textcolor{darkred}{Eff:}| -p2(?from),  p2(?to)

  |\textbf{Pick}|(?p,|\,|?x):
    |\textcolor{darkblue}{Pre:}|  p2(?x),  p1, -p3(?p),  p4(?p,|\,|?x)
    |\textcolor{darkred}{Eff:}|          -p1,  p3(?p), -p4(?p,|\,|?x)

  |\textbf{Drop}|(?p,|\,|?x):
    |\textcolor{darkblue}{Pre:}|  p2(?x), -p1,  p3(?p), -p4(?p,|\,|?x)
    |\textcolor{darkred}{Eff:}|           p1, -p3(?p),  p4(?p,|\,|?x)
        \end{lstlisting}
      \end{minipage}
      & &
      \begin{minipage}[fragile]{\columnwidth}\small
        \noindent Interpretation of learned domain predicates:
        \begin{enumerate}[--]
          \item $p_1$ is true iff agent holds no package (i.e., gripper empty),
          \item $p_2(x)$ is true iff agent is at cell $x$,
          \item $p_3(p)$ is true iff agent holds package $p$,
          \item $p_4(p,x)$ is true iff package $p$ is in cell $x$, and
          \item $p_5(x,y)$ is true if cell $x$ is adjacent to cell $y$. \\
        \end{enumerate}
      \end{minipage}
    \end{tabular}
  \end{tabular}
  \caption{\small \textbf{Top:}
    Labeled state graph $G$ for agent that can move in $1{\times}3$ grid picking  up  and dropping two different packages.
    The graph $G$ has 45 nodes,  assumed to be  black-box states with no  internal structure known  (shown in small circles), 
     and 96 edges, which are  labeled with the actions $\Move$, $\Pick$, and $\Drop$.
    \textbf{Bottom left:}     STRIPS representation $P=\tup{D,I}$ \textbf{learned}  from  $G$:  domain $D$ has  3 action schemas over 5  predicates, and $I$ involves 3 objects (only $D$ is shown).   $P$ is  the \emph{most compact} STRIPS encoding  that yields a graph $G(P)$ that is isomorphic to $G$. 
    The bijection $f$  that underlies  this isomorphism  gives structure to the nodes: black-box node $n$ in $G$ becomes a  planning state
    $f(n)$ in $G(P)$ that assigns a  truth-value to each of the  ground atoms in $P$.
    Action schemas shown in terms of their   preconditions  and effects. Static preconditions that involve predicates whose denotation does not change
    are learned as well.   The domain $D$  learned from the $1{\times}3$ instance with two packages works  for \emph{any}  grid dimensions and \emph{any}
    number of packages.     \textbf{Bottom right:} Interpretation of the learned predicates.
  }
  \label{fig:pkgs3ops:1x3x2}
\end{figure*}

There are, however, \emph{two dimensions} about   knowledge representations that need to be distinguished:
the \emph{representation languages}, such as STRIPS, that are domain-independent,
and  \emph{encodings} in such languages that have traditionally been crafted by hand. The  representation languages 
have been designed with the right goals in mind, including transparency and reuse \citep{mccarthy:generality,pddl:book},
and there is no need to throw them away unless other types of dynamics need to be captured. 
It is the  encodings in the language that constitute the \emph{representation bottleneck,} and it is thus
the  encodings that should be learned from data. 
We will see that this is possible when we  look for the 
most compact representation in the language that explains the data;
a simple a form of Occam's learning that takes advantage
of the powerful and  meaningful inductive bias that results from the
syntax and the semantics of the language.

In the paper we develop this idea of language-based representation learning
and apply it to three related problems: learning general dynamics, learning
general policies, and learning general subgoal structures. The first is about
uncovering the causal structure of a domain; the latter two are about
exploiting it. 
While the paper is original and written for the ``causal representation learning''
audience, the general approach has been developed elsewhere, as described.

\section{Preview}

The top part of Fig.~\ref{fig:pkgs3ops:1x3x2} shows the state graph that results from
a simple, discrete dynamical system that involves an agent that moves on a $1{\times}3$ grid,
that can  pick and drop two different packages, one at a time.
The number of states, 45, results of adding the number of configurations where the agent
holds no package ($27=3^3$) with the number of configurations where the agent holds one
of the packages ($18=2{\times}3^2$).
The number of edges  is  96. 
An  edge $(s,s')$ labeled with an action $a \in \{\Move,\Pick,\Drop\}$ means
that the action $a$ transforms state $s$ into $s'$.
In recent work it has been shown that the internal structure 
of the states can be recovered from state graphs where the states are black-boxes with no known structure.
This is achieved by learning such representations over a first-order causal language able
to represent objects and relations, which are not assumed to be known, and seeking  the
most compact encoding that generates the observed data (i.e., the given state graph).
The language is the modern version of the ``classical'' planning language STRIPS
where a planning instance $P$ is encoded as a pair $P=\tup{D,I}$ with the domain
$D$ being a collection of general action schemas, involving a set of predicates,
and $I$ representing specific instance information. 

Since a planning instance $P$ defines a unique state graph $G(P)$, the learning task
becomes an \emph{inverse problem:} given an observed state graph $G$, find the ``simplest''
planning instance $P=\tup{D,I}$ such that the observed and the generated graphs $G$ and $G(P)$ ``match'' (are isomorphic).
When the graphs match, every \emph{black-box state} (node) in $G$ is mapped into an \emph{structured state} in $G(P)$
given by the  set of ground  atoms obtained from  the  predicates in $D$  and the  set of objects in $I$.
The atoms encode the state variables and their values.

The most compact STRIPS representation $P=\tup{D,I}$ for the state graph shown in Fig.~\ref{fig:pkgs3ops:1x3x2}
has the domain $D$ shown in the bottom part of the figure. The learned domain  consists of 5 
predicates and 3 action schemas that \emph{generalize} to  any instance of the domain,  involving \emph{any} grid dimensions and
\emph{any} number of packages. 
\emph{Objects, relations, and causal structure all emerge from the flat graph shown.
 The learned structure is a result of the data, the target language, and Occam's razor.}

Interestingly,  while the state graph $G$ of a single $1{\times}3$ instance involving two packages yields a domain $D$
that generalizes to any grid size and any number of packages,  smaller training instances yield domains that do not generalize
in the same way. For example, a $1{\times}3$ instance with \emph{one package} yields a model with a unary predicate $p_4(x)$ 
that tracks the position of the unique package and which does not generalize to instances with multiple packages.
Likewise, a $1{\times}2$ instance yields a model with no $p_5(x,y)$ predicate as there is no need then to represent
the topology of the grid (cell adjacency).

\section{Planning Language}

A (classical) planning instance or problem is expressed as a pair $P\,{=}\,\tup{D,I}$ where
the domain $D$ contains a set of action schemas with \emph{preconditions}
and \emph{effects} given by atoms $p(x_1, \ldots, x_k)$ in term of predicates symbols $p$
and variables $x_i$ that are arguments of the action schema,
and the tuple  $I\,{=}\,\tup{Objs, \textit{Init},\textit{Goal}}$ specifies the constants $c_i$ in the \emph{instance} (object names),
and the initial and goal conditions; the latter in terms of \emph{ground atoms} $p(c_1, \ldots, c_k)$
obtained from the domain predicates and the constants \citep{geffner:book,pddl:book}. 
Modern planners usually replace the action schemas by their possible instantiations
where  variables are replaced by constants. Static preconditions are normally
compiled away after pruning the set of instantiations.  A precondition is \emph{static}
when it involves a predicate which  does not appear in an action effect. E.g.,
predicate $p_5$ in Fig.~\ref{fig:pkgs3ops:1x3x2} is a static predicate
which captures the adjacency relation among grid cells. 



A classical problem $P\,{=}\,\tup{D,I}$ encodes a unique state graph $G(P)$ whose nodes
are the states that are reachable in $P$ from the initial state $s_0$. A state $s$ is a collection of ground atoms $q$ from $P$
that encode truth-valuations ($q$ is true in $s$ iff $q \in s$), and the initial state $s_0$
is $Init$. The graph $G(P)$ contains a labeled  edge $(s,a,s')$ if there is a ground instance $a'$
of an action schema $a$ in $D$ that maps the state $s$ into $s'$.
A ground instance $a'$ replaces the schema parameters with constants, and transforms
a state $s$ into $s'$ if the preconditions of $a'$ are true in $s$, the effects of $a'$ are true in $s'$,
and all ground atoms not affected by $a'$ have the same truth value in $s$ and $s'$.

The distinction between the general domain given by $D$ and the specific information given by $I$
in an instance  $P=\tup{D,I}$ is particularly relevant in the  learning setting where learning the structure
of $P$ will mean to learn a general domain $D$ that applies to an infinite number of instances $P'=\tup{D,I'}$
that differ from $P$ in the number of objects, or  in their initial or goal configurations, but not in
the vocabulary (predicates) that captures the structure of the states,  or in the action schemas that capture
the possible state trajectories. In other words, the domain $D$ expresses in a compact way
what is common (invariant) over all these instances.

\section{Learning Planning Models}

The problem of learning general action models from states graphs where states are black boxes
has been formulated as follows \citep{bonet:ecai2020}:

\begin{quoting}
  \noindent\textbf{Learning action models.} Given observed graphs $G_1, \ldots, G_n$, find
  the simplest domain $D$ and instances $P_i=\tup{D,I_i}$ such that the graphs $G_i$ and
  $G(P_i)$ are isomorphic for $i=1, \ldots, n$. 
\end{quoting}

The complexity of a domain is measured in terms of the number and arity of the action schemas and predicates
involved. Once these numbers are bounded, the learning problem becomes a \emph{combinatorial optimization task}
that has been expressed and solved using  SAT and answer set solvers \citep{bonet:ecai2020,ivan:kr2021}.
Variations of this basic problem have also been considered like dealing with noisy and  incomplete traces as opposed to
fully known  graphs \citep{ivan:kr2021}. 

The domains that have been learned in this way include several benchmark domains in planning, from Blocks and Logistics,
to IPC-Grid (a domain similar to Minigrid \citep{babyAI}) and Sokoban. In all cases, one does not only learn the internal structure of the
nodes in the given graphs (i.e., the  atoms encoding the state variables and their values),
but a general domain representation that can be applied to other instances.
This is the result of the strong and meaningful bias that follows from aiming at the most compact
language-based model that matches the data. A crucial part of this is the use of a
\emph{first-order target language} for learning. One action schema represents a potentially infinite set of ground instances.
If rather than looking for the most compact \emph{lifted}  STRIPS representation, we look for the most compact
\emph{propositional} STRIPS representation, very different representations would result.

This is a vanilla solution method, inherently incapable of learning domains dynamics that cannot be expressed
in compact form in lifted STRIPS, but it is a crisp formulation based on general principles and ideas; namely, 
Occam's learning \citep{kearns:book} over an hypothesis space spanned by the target language.
If one wants to learn general dynamics of domains with continuous, exogenous, or non-deterministic changes,
a different target language for learning must be used.
STRIPS is a simple language for modeling deterministic actions, but
there are more expressive planning (action) languages \citep{action-languages,pddl:book},
some of which are used to specify MDPs and POMDPs in compact form
using action schemas, objects, and relations \citep{ppddl,rddl}.

The general idea of language-based representation learning has been used to learn first-order planning
representations from \emph{gray-box states}  represented as objects in 2D grids. 
In this case, the learned  state representations are \emph{grounded}
in the 2D scenes, meaning that there is a 1-to-1 correspondence between scenes and
planning states that generalizes to new scenes \citep{andres:grounded}.

\section{Learning General Policies}

The languages for learning representations can be taken off the shelf
in many cases, but in others, new domain-independent languages may be needed.  
For example, in the Minigrid benchmark \citep{babyAI}, DRL approaches are not after general
dynamic models, but after \emph{general policies}: policies that can deal with \emph{any}
instance of the domain.
What is then a good domain-independent language for representing such policies?
This question has been considered in the area of \emph{generalized planning}, and the
language below follows the one introduced by \citet{bonet:ijcai2018}. 

A general policy $\pi$ for a (possibly infinite) class of instances $\Q$ drawn from a
domain $D$ is given by a set of \emph{policy rules} of the form $C \mapsto E$ where $C$
contains boolean conditions of the form $p$, $\neg p$, $n=0$, or $n > 0$, and $E$
contains effects of the form $p$, $\neg p$, $p?$, $n\mminus$, $n\pplus$, $n?$, over
boolean and numerical \emph{features} $p$ and $n$ that are well-defined over the states
$s$ of any instance from $\Q$.
The action prescribed by the policy $\pi$ in a state $s$ is any action that maps $s$
into a state $s'$ such that the state transition $(s,s')$ satisfies some policy rule
$C \mapsto E$ in $\pi$; namely, $s$ makes $C$ true, and the transition $(s,s')$ makes
the change expressed by $E$ true as well.

For example, with features $\Phi=\{H,p,t,n\}$ for ``holding a package'', ``distances to nearest package and to the target'',
and ``number of undelivered packages'', the following policy solves \emph{any} instance of the domain displayed in  
Fig.~\ref{fig:pkgs3ops:1x3x2} when the goal is to take all the packages, one by one, to a target cell in the grid:
\begin{alignat*}{2}
  &\prule{\neg H,\GT{p}}{\DEC{p},\UNK{t}}\,;             && \text{go to nearest pkg,} \\[.1cm]
  &\prule{\neg H, \EQ{p}}{H}\,;                          && \text{pick it up,} \\[.1cm]
  &\prule{H,\GT{t}}{\DEC{t}}\,;                          && \text{go to target,} \\[.1cm]
  &\prule{H,\GT{n},\EQ{t}}{\neg H, \DEC{n}, \UNK{p}}\,;  &\quad& \text{drop pkg.}
\end{alignat*}
The first rule says to do any action that decreases the distance $p$ to the nearest package ($\DEC{p}$) when not holding
a package and the distance is positive ($\neg H$ and $\GT{p})$, whatever the effect on the distance $t$ to the target ($t?$).
The reading of the other rules is similar with $\DEC{x}$ standing for decrements of feature $x$, and $\UNK{x}$ for any change in $x$.
Features not mentioned in the right-hand side of  a policy rule must keep their values unchanged.

This is a policy written by hand, and the question is how such policies can be learned.
As before, the learning problem has been formulated and solved as a combinatorial optimization problem by creating a large but finite
set of possible boolean and numerical features \emph{from the domain predicates},
using a description logic grammar that captures a decidable fragment of first-order logic,
$\C_2$, where the number of variables is limited to two \citep{description-logics}.
Provided with this pool of features, where each feature is given a cost (the number of grammar rules used to derive it),
the task of learning a general policy becomes \citep{frances:aaai2021}:

\begin{quoting}
  \noindent\textbf{Learning general policies.} Given a known domain $D$, training instances $P_1,\ldots,P_n$,
  over $D$, and a finite pool of domain features ${\cal F}$, each with a cost, find the simplest policy $\pi$
  over $\cal F$ such that $\pi$ solves all $P_i$, $i=1,\ldots,n$. 
\end{quoting}

Once again, the \emph{language} in which policy representations are sought provides a \emph{strongly biased
hypothesis space} where policies that involve few simple features (in terms of the domain
predicates) are preferred. The simplest policies are those that minimize the complexity of the
the features involved. General policies for a number of benchmark planning domains
have been derived in this way and proved to be correct \citep{frances:aaai2021}. More recently, an alternative learning scheme
has been introduced which does \emph{not} require a predefined pool of features \citep{simon:icaps2022,simon:kr2022}.
This is achieved by introducing two variations. First, general value functions $V$ are learned instead of general
policies, so that the resulting policies are those which are greedy in $V$.
Second, the value functions are expressed in terms of graph neural networks (GNNs)
which are known to capture $\C_2$ features \citep{barcelo:gnn,grohe:gnn}.
Interestingly, the resulting policies generalize equally well (100\% generalization
in rich, combinatorial domains) and yield close-to-optimal policies even
in domains where one can prove that there are no general policies that are optimal \citep{blocks-np-hard}.
One point in common with recent deep learning approaches for computing general policies  that
appeal to \emph{causal considerations} \citep{causal:policy1,causal:policy2} is that the learned features are functions of the
domain predicates; namely, the predicates that are used to capture the causal dynamics of the domain.

\section{Learning Subgoal Structure}

The problem of {expressing} and {using} the common subgoal structure of a collection of planning problems
has been important in AI since the 1960s, while the problem of \emph{learning} such
structure has become important in recent RL research 
where useful subgoals are expressed via intrinsic rewards \citep{barto:intrinsic,singh:rewards}.
We are interested in a similar problem but want to learn subgoal structures over a \emph{suitable language.}
The questions, from the perspective of language-based representation learning, are
1)~what is an adequate language for representing subgoal structure, 2)~what is its semantics,
and 3)~how representations over such language can be learned.
A general compact language for representing subgoal structures has been developed recently whose
syntax is the syntax of the general policies considered above.
The change is in the semantics \citep{bonet:aaai2021}.

A \emph{(policy) sketch} is a set of sketch rules $C \mapsto E$
of the same form as {policy rules}, but while policy rules filter 1-step transitions; namely,
when in a state $s$, a 1-step transition to any $s'$ must be selected such that $(s,s')$ satisfies a policy rule,
sketch rules define \emph{subproblems:} when in a state $s$ of an instance $P$, a state $s'$ is to be reached, \emph{not necessarily in one step},
such that the multi-step transition $(s,s')$ satisfies a sketch rule (or $s'$ satisfies the  goal  of $P$).

Sketches \emph{decompose} problems into \emph{subproblems} without prescribing how the
subproblems should be solved (going from $s$ to $s'$). One is interested, however, in sketches
that yield subproblems that can be solved efficiently, in low polynomial time
(in the number of problem variables), and this is guaranteed when subproblems
have \emph{bounded width} \citep{nir:ecai2012}. This observation led  to the following
formulation for learning sketches, where the notation $P[R]$ is used to refer to the
collection of subproblems defined by the sketch $R$ over states $s$ that are reachable
in the instance $P$ \citep{dominik:icaps2022}:

\begin{quoting}
 \noindent\textbf{Learning general sketches.} Given a known domain $D$, training instances
  $P_1,\ldots,P_n$ and a non-negative integer $k$, find the simplest sketch $R$ over a pool
  of domain features $\cal F$ such that 1)~the collection of subproblems induced by $R$ on
  each instance $P_i$, $P_i[R]$, have \emph{width bounded}  by $k$, and 2)~the sketch $R$ is acyclic in $P_i$,
  $i=1,\ldots,n$.
\end{quoting}

The complexity of a sketch is given by the complexity of the features involved, and a sketch
is acyclic in $P$ if the transitions $(s,s')$ in $P$ that satisfy sketch rules do not form a cycle.
The learning problem becomes a combinatorial optimization problem modeled and solved using the
answer set programming system Clingo \citep{clingo}.

The learned sketches are not aimed at representing the general causal structure of the domain, but
at exploiting it. Indeed, by learning to decompose problems into subproblems of bounded width,
the problems can be solved in polynomial time using  a general algorithm ($\siwR$) that takes the
sketch into account \citep{bonet:aaai2021,dominik:kr2021}.
Simple examples of learned sketches follow. 

A width-2 sketch $R_1$ for the problem above, where packages need to be delivered to a target cell,
one by one, involves the feature $n$ which tracks the number of packages not yet delivered, and
is given by a single rule:
\begin{alignat*}{2}
  &R_1: \{ \prule{\GT{n}}{n\mminus} \} \,.
\end{alignat*}
The  sketch expresses a decomposition that for states $s$ where $\GT{n}$, states $s'$ should be
reached where the value of $n$ is lower than in $s$. One can show that the resulting subproblems
have width bounded by $2$ and thus  can  be solved by running the IW(2) algorithm \citep{nir:ecai2012}.

A width-1 sketch $R_2$ that involves the features $n$ and $H$ ($H$ is ``holding a package'')
is given instead by two rules:
\begin{alignat*}{2}
  &R_2: \{ \prule{\neg H}{H} \, , \, \prule{\GT{n},H}{n\mminus,\neg H} \} \,.
\end{alignat*}
The first rule on the left says that if not holding a package, one such package should be picked up,
while the second on the right says that if holding a package, it should be delivered.
The rules do not express  policies but subgoals to be achieved.
In this case, the subproblems have all width $1$ meaning that they can be solved
in linear time by running the IW(1) algorithm.

\section{Related Work}

\textbf{Structured causal models.} SCMs are the standard language for encoding and studying causal models \citep{pearl:causal}.
Yet, planning and action languages also encode causal models;
i.e., they can accommodate observations,
interventions, and counterfactuals; all  3-levels of Pearl's Causation Ladder \citep{pearl:bow}).\footnote{For doing this, 
  uncertainty about the initial situation must be represented,  and action precondition and effects in STRIPS
  need to be replaced by conditional effects, a standard feature of modern planning languages \citep{pddl:book}.
  Then, the atoms at time $0$ and the actions at each time step are the exogenous variables,
  and  the atoms at time $t+1$ for $t > 0$ are the endogenous variables. The truth value of the latter
  is a  function of the value of atoms  at time $t$ and the action at $t$. Planning languages just provide a  way for 
  defining the atoms (i.e., the Boolean variables of interest), the functions, and in certain cases, the uncertainty
  about the initial  situation, in a compact and reusable form.}
SCMs enrich the flat language of probabilities but are not rich enough to express interventions that
do not commute unless the variables are indexed with time in the style of dynamic bayesian networks \citep{dbns}.
Some planning languages for modeling MDPs and POMDPs are variations of STRIPS and PDDL \citep{ppddl}; others,
of dynamic bayesian networks \citep{rddl}. More critically for representation learning,
SCMs are propositional and not first-order, which precludes learning compact representations in settings
that involve objects. The problem of learning SCMs when the variables are known has been addressed as a combinatorial
optimization problem \citep{causal:asp1,causal:asp2}, but it is more subtle and lacks a general formulation
when the variables are not known \citep{bernhard:causal,bernhard:causal2}. We have shown that the use of
first-order causal languages along with a preference for compact representations uncovers objects, relations,
and causes from state graphs made up of black-box states. 

\medskip
\noindent\textbf{Neuro-symbolic AI.} The use of formal representation languages in a learning setting is
common of neuro-symbolic approaches where the languages are used to encode background knowledge
\citep{luciano:ltn,luc:neuro,asp:nnet,kersting:aspnn}. In language-based representation learning,
on the other hand, no background knowledge is assumed, and the language provides the general
structure over which the representations are learned \citep{geffner:target}.

\medskip
\noindent\textbf{Language-based representation learning.}. Many works learn representations over languages
without using background knowledge \citep{locm,konidaris:jair,asai:prop1,making-sense,making-sense-raw,konidaris:2022}.
In some cases, these are programming or visual languages \citep{josh:programming,littman:visual};
in others, they provide the intuitions that underlie the design of  deep network architectures  aimed at capturing
production systems, action schemas, and other classes of first-order formulas \citep{rim,scoff,bengio:rules,shanahan:predinet}.
Often though,  deep learning approaches are developed without providing  a crisp characterization
of \emph{what} representations are to be learned that is independent of \emph{how} they are learned.
We have shown that action schemas and predicates can be learned using combinatorial solvers but
nothing prevents the use of deep learning engines instead.
General policies were learned using combinatorial solvers \citep{frances:aaai2021},
and then using  deep learning \citep{simon:icaps2022,simon:kr2022}.


\medskip
\noindent\textbf{General policies, subgoals, and intrinsic rewards.}
Some approaches for learning general policies make use of languages for representing policies \citep{khardon:generalized,martin:kr2000,fern:general};
but most of those based on deep learning, do not \citep{sid:general,mausam:general,sylvie:general}.
Subgoals in planning have been expressed in terms of hierarchical task networks \citep{htn:planning} but a recent language \citep{bonet:aaai2021}
supports more compact representations that facilitates learning of both policies and sketches \citep{frances:aaai2021,dominik:icaps2022}.
In RL, subgoals are associated with states of intrinsic reward (stepping stones to sparse states of extrinsic reward) \citep{singh:rewards},
but no language or principles have been developed for expressing or learning them, and the focus is on performance improvement
that is less informative and crisp.

\section{Discussion}

The language of structured causal models has been fundamental to model and to understand causality,
but the language itself is insufficient to model the world. Objects and schemas are required
to represent the dynamics of the world in a compact way, and this ability is a prerequisite for
learning it.\footnote{The old AI saying that ``you can't learn what you can't represent'',
  still contains a grain of truth in the age of deep learning.}
We have advocated the use of a broader class of domain-independent languages for learning causal
representations, including those required to model dynamics, policies, and subgoals. Some of these languages can
be taken off-the-shelf; others have to be designed. It is the languages and the preference for compact representations over them
that structure the world and uncover objects, relations, and causes. The languages considered for modeling and learning
system dynamics, leave many important aspects aside, including continuous, exogenous, and non-deterministic change.
For addressing such aspects, richer languages and more powerful learning methods, possibly based on deep learning,
are needed. Learning reusable and meaningful language-based dynamic models of, say, the Atari games from the screen
pixels alone, is still an open challenge  that we think could be addressed in the near future.

\subsection*{Acknowledgments}

This research was partially supported by the European Research Council (ERC), Grant No.\ 885107, and by project TAILOR,
Grant No.\ 952215, both funded by the EU Horizon research and innovation programme.
This work was partially supported by the Wallenberg AI, Autonomous Systems and Software Program (WASP) funded by the Knut and Alice Wallenberg Foundation.

\bibliographystyle{aaai22}
\bibliography{control}


\end{document}